\documentclass[conference]{IEEEtran}
\IEEEoverridecommandlockouts
\usepackage{cite}
\usepackage{amsmath,amssymb,amsfonts}
\usepackage{algorithmic}
\usepackage{graphicx}
\usepackage{textcomp}
\usepackage{xcolor}

\usepackage{hyperref}
\usepackage{bbding}
\usepackage{multirow}
\usepackage{multicol}
\usepackage{dblfloatfix}
\usepackage{booktabs}
\usepackage[T1]{fontenc}
\def\BibTeX{{\rm B\kern-.05em{\sc i\kern-.025em b}\kern-.08em
    T\kern-.1667em\lower.7ex\hbox{E}\kern-.125emX}}

\makeatletter
\newcommand{\linebreakand}{%
  \end{@IEEEauthorhalign}
  \hfill\mbox{}\par
  \mbox{}\hfill\begin{@IEEEauthorhalign}
}
\makeatother

\begin{document}

\title{SwapTalk: Audio-Driven Talking Face Generation \\ with One-Shot Customization in Latent Space\thanks{\Envelope Corresponding author. \quad $^*$\ Equal contribution.}
}

\author{\IEEEauthorblockN{Zeren Zhang$^{*}$}
\IEEEauthorblockA{\textit{Peking University} \\
Eric\_Zhang@stu.pku.edu.cn}
\and
\IEEEauthorblockN{Haibo Qin$^{*}$}
\IEEEauthorblockA{\textit{Youdao AI} \\
qinhaibo@rd.netease.com}
\and
\IEEEauthorblockN{Jiayu Huang}
\IEEEauthorblockA{\textit{Youdao AI} \\
huangjy04@rd.netease.com}
\and
\IEEEauthorblockN{Yixin Li}
\IEEEauthorblockA{\textit{Youdao AI} \\
yixinli@rd.netease.com}
\linebreakand
\IEEEauthorblockN{Hui Lin}
\IEEEauthorblockA{\textit{Youdao AI} \\
linhui@rd.netease.com}
\and
\IEEEauthorblockN{Yitao Duan}
\IEEEauthorblockA{\textit{Youdao AI} \\
duan@rd.netease.com}
\and
\IEEEauthorblockN{Jinwen Ma\Envelope}
\IEEEauthorblockA{\textit{Peking University} \\
jwma@math.pku.edu.cn}
}

\maketitle

\begin{abstract}
Combining face swapping with lip synchronization technology offers a cost-effective solution for customized talking face generation. However, directly cascading existing models together tends to introduce significant interference between tasks and reduce video clarity because the interaction space is limited to the low-level semantic RGB space. To address this issue, we propose an innovative unified framework, SwapTalk, which accomplishes both face swapping and lip synchronization tasks in the same latent space. Referring to recent work on face generation, we choose the VQ-embedding space due to its excellent editability and fidelity performance. To enhance the framework's generalization capabilities for unseen identities, we incorporate identity loss during the training of the face swapping module. Additionally, we introduce expert discriminator supervision within the latent space during the training of the lip synchronization module to elevate synchronization quality. In the evaluation phase, previous studies primarily focused on the self-reconstruction of lip movements in synchronous audio-visual videos. To better approximate real-world applications, we expand the evaluation scope to asynchronous audio-video scenarios. Furthermore, we introduce a novel identity consistency metric to more comprehensively assess the identity consistency over time series in generated facial videos. Experimental results on the HDTF demonstrate that our method significantly surpasses existing techniques in video quality, lip synchronization accuracy, face swapping fidelity, and identity consistency. Our demo is available at \url{http://swaptalk.cc}.
\end{abstract}

% \begin{IEEEkeywords}
% Face swapping, Talking face generation, Audio driven animation
% \end{IEEEkeywords}
\section{Introduction}
In recent years, significant technological advancements have been made in the field of audio-driven talking face generation technologies~\cite{Ki_2023_ICCV, guan2023stylesync, gupta2023towards, wang2023lipformer, stypulkowski2024diffused} for virtual digital humans. However, generating a lip-synchronized talking face video from a user's customized portrait remains challenging. In this context, combining face swapping with lip synchronization (lip-sync) technology provides a cost-effective and practical solution.

The most intuitive way is serially cascading the face swapping model and the lip-sync model to meet the demands of this customization task. However, this straightforward cascade approach presents significant mutual interference issues. If lip-sync is performed before face swapping, the face swapping model may fail to precisely retain the details of the lips, thereby compromising the accuracy of audio-visual synchronization. Conversely, if face swapping is conducted before lip-sync, while the synchronization of lip movements might be effectively achieved, it could degrade the quality of the face swapping and the overall consistency of the generated video. We speculate that this issue is primarily due to the direct concatenation of the two models in the RGB space. Given that the RGB space contains a wealth of facial details and low-level semantics, its editability and decoupling properties are limited, making it unsuitable as a basis for interaction between the two models. Moreover, directly cascading existing models may result in reduced clarity. Although image reconstruction techniques can partially improve clarity, they may also compromise lip synchronization accuracy and introduce errors in face swapping details.

\begin{figure}[t]
    \centering
    \includegraphics[width=9cm]{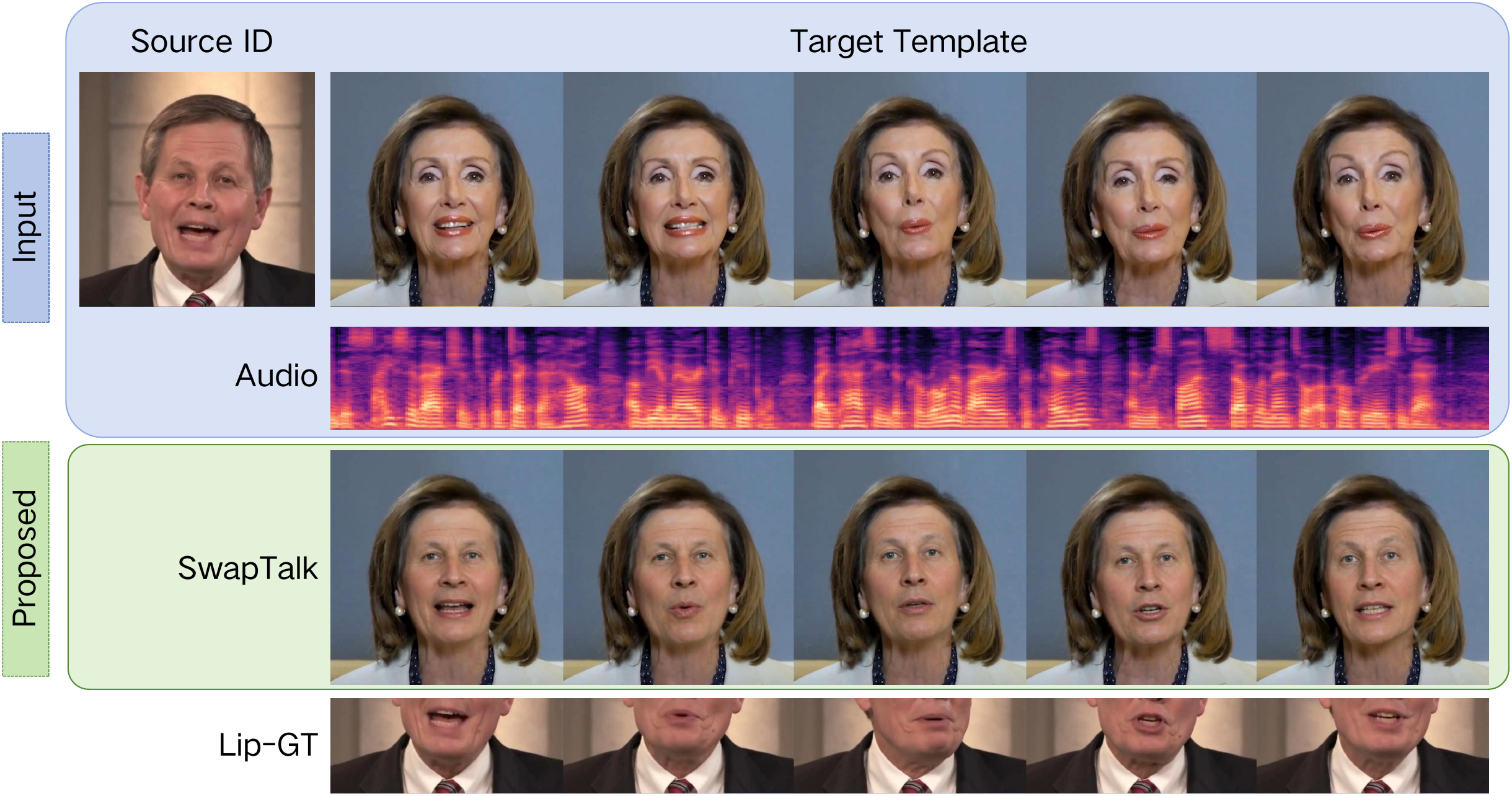}
    \caption{Our model is capable of transferring the facial region of a user-defined personalized avatar (source ID) onto a specified target template, while also accommodating lip shape deformations to ensure that the lip movements in the generated video are synchronized with the user-specified audio content.}
    \label{fig:first-page}
\end{figure}

To address the issues previously discussed, we propose an innovative, unified framework called SwapTalk. This framework manages face swapping and lip-sync tasks within a shared latent space, expecting to enhance the accuracy of both tasks and improve overall consistency. Inspired by~\cite{gupta2023towards, Rombach_2022_CVPR}, our framework is built on the decoupled and high-fidelity VQ-embedding space of the pre-trained Vector Quantized Generative Adversarial Network~\cite{Esser_2021_CVPR} (VQGAN). Learning both tasks in this compact latent space reduces the computational cost for the face swapping and lip-sync modules. Additionally, it simplifies the model's learning process by leaving high-resolution image generation to the pre-trained VQGAN.

%To address the issues previously discussed, we propose SwapTalk, an innovative framework that manages face swapping and lip-sync tasks within a single, shared latent space. Inspired by recent advances in generative networks (as discussed in Gupta et al., 2023, and Rombach et al., 2022), SwapTalk leverages the decoupled, high-fidelity VQ-embedding space of the pre-trained Vector Quantized Generative Adversarial Network (VQGAN) (Esser et al., 2021). This integration aims to enhance the accuracy of both tasks and ensure greater overall consistency. By operating in this compact latent space, SwapTalk not only reduces computational demands but also simplifies the learning process by delegating high-resolution image generation to the pre-trained VQGAN.

In developing our modules, we have redesigned the face swapping module using the Transformer~\cite{vaswani2017attention} and the lip-sync module using the UNet from~\cite{Rombach_2022_CVPR}, both utilizing a VQ-embedding space. During the training phase, these modules are trained independently. We employ an identity loss during the face swapping module's training, which greatly improves the model's ability to handle unseen identities. Furthermore, we enhance the lip-sync module with lip-sync expert supervision within the VQ-embedding space, which increases the accuracy of lip synchronization. During the inference phase, we thoroughly assess the sequence of applying these modules and find that performing face swapping before lip-sync within the VQ-embedding space yields the most effective results, forming the foundation of our SwapTalk framework. For visual results, please see Figure~\ref{fig:first-page}.

During the evaluation phase, we find that the comparison of previous lip-sync models mainly focuses on the task of self-reconstruction of lip movements in audio-visual synchronous videos, which does not sufficiently meet the requirements of practical applications. Therefore, we expanded the scope of testing to include evaluations in actual situations where audio and video are not synchronized, to better fit real-world application scenarios. Considering that our framework involves both face swapping and lip-syncing tasks, we designed two testing scenarios: self-driven (where the audio source matches the original identity in the video) and cross-driven (where the audio source differs from the original identity). Testing under these two scenarios can more accurately reflect the performance of models under various application conditions. In addition, to evaluate facial identity consistency over time series in videos, we introduce a novel identity consistency metric. This metric aims to comprehensively assess the performance of existing facial video generation models, especially their effectiveness in maintaining identity consistency.

Based on the discussions above, our main contributions can be summarized as follows:
\begin{itemize}
    \item  We propose a unified framework that completes face swapping and lip-sync tasks within a semantically rich and decoupled VQ-embedding space, simultaneously achieving lip synchronization and preserving both ID appearance and overall consistency.
    \item We utilize the identity loss during the training process of the pre-trained VQGAN and the face swapping module to enhance the generalization of unseen identities and consistency across time series. Additionally, we employ a lip-sync expert within the VQ-embedding space to improve lip synchronization accuracy.
    \item We point out issues with lip synchronization evaluation in prior research and propose the use of an audio-visual un-synchronized setting to more accurately assess lip synchronization in realistic scenarios. Furthermore, we introduce a metric designed to reasonably evaluate facial identity consistency within videos.
\end{itemize}

\section{Related Work}
Our research primarily involves two domains: \textbf{Face Swapping} and \textbf{Talking Face Generation}.

\paragraph{Face Swapping}
Face swapping is a task to synthesize an image with the identity of the source image while preserving the target image’s attributes (e.g., expression, pose, and shape). We typically categorize existing face-swapping models into 3D-based and GAN-based methods. 3D-based approaches~\cite{blanz2004exchanging, nirkin2018face} commonly employ 3D Morphable Models~\cite{blanz2023morphable} (3DMM) to incorporate structural prior information and directly replace target faces with synthesized ones through blending or warping modules. However, these methods generally require manual intervention and can produce unrealistic effects under complex lighting conditions. Recently, GAN-based face swapping~\cite{li2019faceshifter, li2020advancing, chen2020simswap, xu2022styleswap, kim2022smooth} has emerged as a popular area of research. These methods typically employ an encoder-decoder architecture that injects identity information from the source face into the GAN's hidden space and decoder's feature layers to guide facial region replacement on the target face. However, these approaches commonly involve multiple loss functions and necessitate careful tuning of their hyper-parameters to achieve balance. Additionally, they often struggle to accurately align facial features when the source and target faces differ significantly in shape. To address these limitations, recent approaches~\cite{li2021faceinpainter, wang2021hififace, xu2022high, xu2022region} have integrated 3DMM features and facial landmarks as additional information injected into GANs to achieve more accurate face swapping results. Besides, DiffSwap~\cite{zhao2023diffswap} employs a denoising diffusion probabilistic model (DDPM) to obtain high-fidelity swap faces. However, these models remain unstable and difficult to optimize owing to costly computational demands in a large number of denoising steps.

\paragraph{Talking Face Generation}

Talking face generation has been an active area of research in recent years. Our work focuses on audio-driven lip synchronization, which involves deforming the mouth area in the target video while keeping other information such as posture unchanged. 
Wav2Lip~\cite{prajwal2020lip} is one of the most classic works in this field. It utilizes a standard encoder-decoder architecture to directly learn from RGB space, supervised by a lip-sync expert for lip shape generation. However, this model faces a trade-off between video clarity and lip synchronization accuracy~\cite{gupta2023towards}, resulting in a low face resolution of only $96\times 96$ pixels. 
An effective solution is to leave the high-fidelity image generation to the pre-trained GANs while achieving lip synchronization through manipulation of the GAN's hidden space.
Recent studies, such as~\cite{alghamdi2022talking, guan2023stylesync, Ki_2023_ICCV} have demonstrated practical applications of this technique by combining audio information to modify the $\mathcal{W}^{+}$ space of StyleGAN~\cite{karras2020analyzing}. Moreover, StyleHEAT~\cite{yin2022styleheat} also utilizes 3DMM parameters as additional information to generate flow fields that guide the decoding and generation process of StyleGAN. 
In addition to StyleGAN, methods that directly manipulate the latent space of VQGAN~\cite{gupta2023towards, wang2023lipformer} also offer a new perspective for generating high-definition lip synchronization videos.

WAVSYNCSWAP~\cite{bao2023wavsyncswap} represents the pioneering model that accomplishes lip synchronization with face swapping within a unified framework, though it still lags in video clarity and consistency of character identities. Detailed comparisons will be presented in the experimental section.
\section{Approach}
\begin{figure*}[htbp]
    \centering
    \includegraphics[width=18cm]{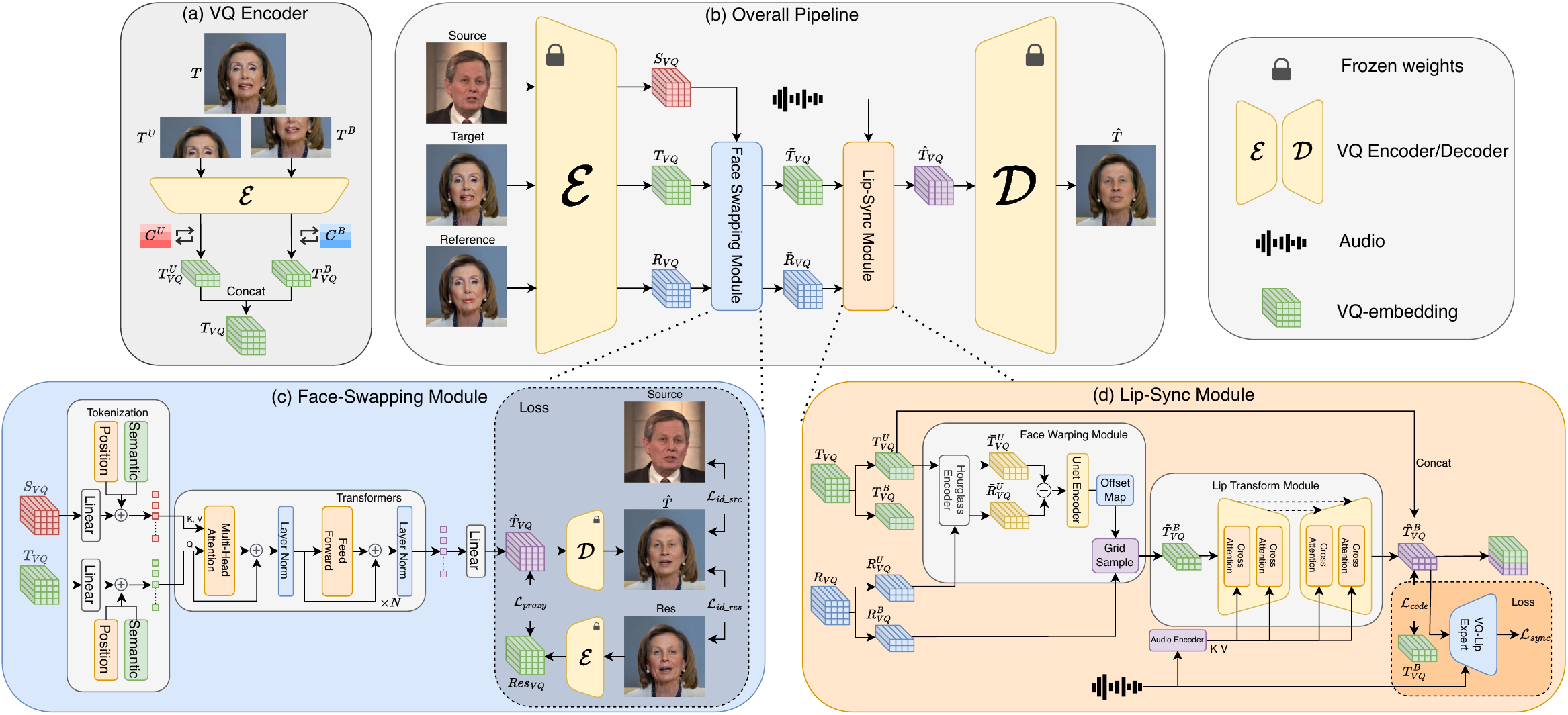}
    \caption{(a) Details of the encoding process of the VQ Encoder. (b) The overall framework of our proposed method. The facial image is first encoded into the VQ-embedding space. Then, the face swapping module (c) and the lip-sync module (d) handle face swapping and lip synchronization, respectively. Finally, the VQ Decoder converts the output back into RGB space, producing a customized talking face video.}
    \label{fig:overall_pipeline}
\end{figure*}

We pre-train a VQGAN~\cite{Esser_2021_CVPR} on a large dataset of high-definition facial images to serve as the foundational model for our framework. After this phase, we freeze the base model and develop the face swapping and lip-sync modules within the established VQ-embedding space. Figure~\ref{fig:overall_pipeline}(b) illustrates the overall pipeline of our proposed framework. It requires three facial images---source, target, and reference---as well as an audio clip for input. Initially, the pre-trained VQ Encoder projects these facial images into the VQ-embedding space, represented as $S_{VQ}$, $T_{VQ}$, and $R_{VQ}$. The face swapping module then transfers the source face onto the target and reference faces, generating $\tilde{T}_{VQ}$ and $\tilde{R}_{VQ}$ respectively. Subsequently, the lip-sync module uses $\tilde{T}_{VQ}$ as the pose reference and $\tilde{R}_{VQ}$ as the lip texture reference, integrating the audio input to predict the lip-synced VQ-embedding, denoted as $\hat{T}_{VQ}$. Finally, the VQGAN decoder converts $\hat{T}_{VQ}$ back into the RGB space, yielding the final prediction $\hat{T}$. Detailed explanations of the pre-trained VQGAN and the architecture of our face swapping and lip-sync modules are provided in Sections~\ref{subsec: pre-trained vqgan}, ~\ref{subsec: swap module}, and ~\ref{subsec: lip-sync module} respectively.

\subsection{Pre-trained VQGAN} \label{subsec: pre-trained vqgan}
Following the practice in~\cite{wang2023lipformer}, we separately use two codebooks $C^{U}$ and $C^{B}$ to encode the upper and lower halves of the face. We expect $C^{U}$ to learn the head pose from the upper facial region, and $C^{B}$ to learn the lip movement from the lower portion of the face. Each codebook contains 2048 codes, and each code is a 256-dimensional vector. As shown in Figure~\ref{fig:overall_pipeline}(a), the input facial image $T$ is pre-processed to a uniform size of $320\times 320\times 3$ and divided into upper and lower halves, $T^{U}, T^{B}$. The encoding process by VQ Encoder is as follows:
\begin{equation}
\begin{aligned}
T^{U}_{VQ} &= \text{Quant}(\text{Enc}(T^{U}), C^{U})\\
T^{B}_{VQ} &= \text{Quant}(\text{Enc}(T^{B}), C^{B})\\
T_{VQ} &= \text{Cat}([T^{U}_{VQ}, T^{B}_{VQ}]),
\end{aligned}
\end{equation}
where $T^{U, \text{raw}}_{VQ}:=\text{Enc}(T^{U}), T^{B, \text{raw}}_{VQ}:=\text{Enc}(T^{B})$ are the initial encoding of the upper and lower halves obtained through the VQ Encoder. The $\text{Quant}(T^{*, \text{raw}}_{VQ}, C^{*})$ process involves looking up the nearest VQ-code in the corresponding codebook to obtain the VQ-embedding. This encoding process can be simplified as follows:
\begin{equation}
    T_{VQ} = \text{Enc}_{VQ}(T).
\end{equation}
The VQ-embedding can be decoded by the VQ Decoder to reconstruct the facial image, $\hat{T} = \text{Dec}_{VQ}(T_{VQ})$. During the training process of VQGAN, the model not only incorporates the conventional GAN loss $\mathcal{L}_{GAN}$, but also includes a reconstruction loss which is composed of both pixel-level $L_{1}$ loss and perceptual losses~\cite{wang2018high}:
\begin{equation}
    \mathcal{L}_{rec} = \|T - \hat{T} \|_{1} + \sum_{i=1}^{M} \|f_{vgg}^{(i)}(T) -  f_{vgg}^{(i)}(\hat{T}) \|_{1}.
\end{equation}
Here, $f_{vgg}^{(i)}(\cdot)$ represents the features from the $i$-th layer of the VGG network. Besides, the VQ-quantization loss is also included:
\begin{equation}
    \mathcal{L}_{VQ} = \|\text{sg}[T_{VQ}] - T^{\text{raw}}_{VQ}\|_{1} + \|T_{VQ} - \text{sg}[T^{\text{raw}}_{VQ}]\|_{1},
\end{equation}
where $T^{\text{raw}}_{VQ} = \text{Cat}[T^{U, \text{raw}}_{VQ}, T^{B, \text{raw}}_{VQ}]$ is the concatenated encoding before quantization, and $\text{sg}[\cdot]$ denotes the stop-gradient operator. Furthermore, to enhance the model's ability to maintain identity during the reconstruction process, we also introduce an identity loss calculated by Arcface~\cite{deng2019arcface} during the pre-training process:
\begin{equation}
    \mathcal{L}_{id} =\|\text{Arcface}(T) - \text{Arcface}(\hat{T})\|_{1}.
\end{equation}
The final pre-training loss is a weighted sum of these four losses:
\begin{equation}
    \mathcal{L}_{\text{pre-trained}} = \mathcal{L}_{GAN} + \lambda_{rec}\cdot\mathcal{L}_{rec} + \lambda_{VQ}\cdot\mathcal{L}_{VQ} + \lambda_{id}\cdot\mathcal{L}_{id},
\end{equation}
where \(\lambda_{rec}, \lambda_{VQ}, \lambda_{id}\) are the weight coefficients.

\subsection{Face Swapping Module} \label{subsec: swap module}
Figure~\ref{fig:overall_pipeline}(c) depicts our face swapping module configured within the VQ-embedding space, consisting of a tokenization module and several Transformer Encoders. The module processes the latent representations of the input source and target faces, denoted as $S_{VQ}$ and $T_{VQ}$ respectively, within the VQ-embedding space. The tokenization module first applies a linear layer to adjust these representations to the input dimension size $d_{model}$, compatible with the Transformers. To further define the position and origin of these tokens, we introduce position and semantic embeddings. In our process, we utilize the transformed target tokens as queries ($Q$), and the source tokens as keys ($K$) and values ($V$). This setup facilitates information exchange and integration via a cross-attention mechanism.
\begin{equation}
    \text{Attention}(Q, K, V) = \text{softmax}\left(\frac{QK^T}{\sqrt{d_k}}\right)V
\end{equation}
After multiple layers of interactions within Transformers, the output is then transformed back into the VQ-embedding space through another simple linear layer, forming the predicted result of the face swapping module (denoted as $\hat{T}_{VQ}$).

During training, our module first learns from a proxy face swapping model. Assuming the proxy model's face swapping result is $Res$, we encode it into the VQ-embedding space using a pre-trained VQ Encoder to obtain $Res_{VQ} = Enc_{VQ}(Res)$. This encoded result serves as a learning target for our face swapping module:
\begin{equation}
    \mathcal{L}_{proxy} = \|Res_{VQ} - \hat{T}_{VQ} \|_{1}.
\end{equation}
Besides, We observe that learning face swapping solely from the proxy model does not effectively generalize to unseen identities (this discussion is in Section 4.4). Consequently, we remap the predicted $\hat{T}_{VQ}$ to RGB space using a pre-trained VQ Decoder (represented as $\hat{T}$). Then we introduce the identity loss between $\hat{T}$ and the source face ($S$), as well as between $\hat{T}$ and the proxy model's swapping result, to enhance identity consistency in the generated videos: 
\begin{equation}
\begin{aligned}
\mathcal{L}_{id\_src} &= \|\text{Arcface}(S) - \text{Arcface}(\hat{T}) \|_{1}, \\
\mathcal{L}_{id\_res} &= \|\text{Arcface}(Res) - \text{Arcface}(\hat{T}) \|_{1}.
\end{aligned}
\end{equation}
Finally, the loss function for our face swapping module is a weighted sum of the proxy loss and the two identity losses:
\begin{equation}
\mathcal{L}_{face-swapping} = \mathcal{L}_{proxy} + \lambda_{src} \cdot \mathcal{L}_{id\_src} + \lambda_{res} \cdot \mathcal{L}_{id\_res}.
\end{equation}

\subsection{Lip-Sync Module} \label{subsec: lip-sync module}
The design of the lip-sync module, as shown in Figure~\ref{fig:overall_pipeline}(d), is composed of two sub-modules, face warping and lip transform, which respectively handle pose transfer and lip shape modification after the input of target and reference VQ-embedding. Initially, the input is split into representations of the upper and lower halves of the face:
\begin{equation}
\begin{aligned}
T_{VQ} := \text{Cat}([T^{U}_{VQ}, T^{B}_{VQ}]) \quad R_{VQ} := \text{Cat}([R^{U}_{VQ}, R^{B}_{VQ}]).
\end{aligned} 
\end{equation}
The face warping module calculates the difference between $T^{U}_{VQ}$ and $R^{U}_{VQ}$ and applies the offset map and grid sample to the lower part of the reference face $R^{B}_{VQ}$, resulting in an estimate of the lower part of the target face, $\tilde{T}^{B}_{VQ}$, which aligns the pose differences. Subsequently, the lip transform module, guided by the input audio, deforms $\tilde{T}^{B}_{VQ}$ in the mouth region, predicting the VQ-embedding $\hat{T}^{B}_{VQ}$ synchronized with the speech. The lip transform module utilizes an UNet structure from~\cite{Rombach_2022_CVPR} (denoted as $F_{ldm\_unet}(\cdot, \cdot)$), which can incorporate conditional inputs. The conditional inputs are injected into the backbone network structure through cross attention, influencing the $8\times$ and $4\times$ down-sampling feature layers of $\tilde{T}^{B}_{VQ}$. The input audio is pre-processed by HuBERT~\cite{hsu2021hubert} (denoted as $Enc_{H}(\cdot)$) to extract speech features, which are then injected as conditions into the lip transform module to obtain an estimate of the lower part of the face's VQ-Embedding synchronized with the speech:
\begin{equation}
\hat{T}^{B}_{VQ} = F_{ldm\_unet}(\tilde{T}^{B}_{VQ}, Enc_{H}(A)).
\end{equation}
The complete prediction result is obtained by concatenating $T^{U}_{VQ}$:
\begin{equation}
    \hat{T}_{VQ} = \text{Cat}([T^{U}_{VQ}, \hat{T}^{B}_{VQ}]).
\end{equation}
During training, the target and reference inputs come from different frames of the same video, while the audio remains synchronized with the target. At this stage, the ground truth for the predicted $\hat{T}^{B}_{VQ}$ of the lip-sync module should be $T^{B}_{VQ}$. The $L_{1}$ distance between these two VQ-embeddings is used as the encoding loss in the latent space:
\begin{equation}
   \mathcal{L}_{code} = \|T^{B}_{VQ} - \hat{T}^{B}_{VQ}\|_{1}.
\end{equation}
In addition, similar to~\cite{gupta2023towards}, we additionally train a lip-sync expert in the VQ-embedding space to supervise lip synchronization. This module takes the lower face's VQ-embedding and audio(denoted as $A$) as inputs, encoding them separately using two encoders ($Enc_{VQ}^{Expert}(\cdot)$ and $Enc_{A}^{Expert}(\cdot)$), and computes the cosine similarity between the feature layers as the lip synchronization score. The lip synchronization loss at this stage is defined as:
\begin{equation}
    \mathcal{L}_{sync} = -\log(cos(Enc_{VQ}^{Expert}(\hat{T}^{B}), Enc_{A}^{Expert}(A))).
\end{equation}
The total loss of our lip-sync module is:
\begin{equation}
   \mathcal{L}_{Lip-Sync} = \mathcal{L}_{code} + \lambda_{sync}\cdot\mathcal{L}_{sync}.
\end{equation}
In the inference phase, the approach differs based on the task type. For self-driven tasks—those involving synchronized input audio and target video—we initialize the $R_{VQ}$ with the VQ-embedding from the target video's first frame. This strategy prevents lip shape leakage issues. Conversely, in cross-driven tasks, where there is a lack of synchronization between the input audio and target video, $R_{VQ}$ is assigned the same value as $T_{VQ}$. Additionally, we bypass the face warping module during inference by directly setting $\tilde{T}^{B}_{VQ}$ equals to $T^{B}_{VQ}$. The alignment of audio with the lip's movements is then achieved through the lip transform module.

\subsection{Consistency Metric}
In the task of talking face generation, maintaining identity consistency in generated videos is crucial. Therefore, we propose an innovative metric to measure identity consistency. For a given generated video clip $\{T^{(i)}\}_{i=1}^{T}$, we utilize a pre-trained face recognition network---distinct from the Arcface architecture previously used in training VQGAN and face swapping---to extract features $F_{i} = Enc_{face}(T^{(i)})$. We then calculate the average cosine similarity between these features and those of the source face, defining this measure as the video’s identity consistency score:
\begin{equation}
    \text{Consistency} = \frac{1}{T}\sum_{i=1}^{T} \cos(Enc_{face}(T^{(i)}), Enc_{face}(S)).
\end{equation}

\section{Experiment}

\subsection{Dataset}

\begin{table*}[htbp]
    \centering
    \renewcommand{\arraystretch}{0.8}
    \caption{Quantitative results of our model with various baseline models on the HDTF test set for both self-driven and cross-driven settings. \\ ${\dagger}$ indicates the results calculated using the demo video of the WAVSYNCSWAP.}
    \begin{tabular}{c|c|ccc|cc|cc}
        \toprule
        Task Type & Methods  & FID$\downarrow$ & SSIM$\uparrow$ & CPBD$\uparrow$ & LMD$\downarrow$ & LSE-C$\uparrow$ & ID Retrieve$\uparrow$ & Consistency$\uparrow$\\
        \midrule
        \multirow{10}{*}{Self-Driven} & Ground Truth & - & - & 0.536 & - & 8.37 & - & - \\
         & Wav2Lip~\cite{prajwal2020lip} & 28.5 & 0.813 & 0.425 & 1.959 & \textbf{9.50} & - & - \\
         & Wav2Lip-Restore & 15.1 & 0.904 & 0.535 & 1.624 & 9.04 & - & - \\
         & Sync-Swap & 12.4 & 0.904 & 0.484 & 1.481 & 7.49 & 82.9 & 76.00 \\
         & Sync-Swap-Restore & 12.7 & 0.885 & \textbf{0.539} & 1.483 & 7.25 & 81.4 & 75.57 \\
         & Swap-Sync & 11.7 & 0.906 & 0.482 & 1.384 & 8.89 & 80.3 & 74.89 \\
         & Swap-Restore-Sync & 12.5 & 0.890 & 0.533 & 1.386 & 8.84 & 79.3 & 74.38 \\
         & WAVSYNCSWAP~\cite{bao2023wavsyncswap} & 49.9 & 0.738$^{\dagger}$ & 0.470 & 3.161 & 9.09 & 85.7 & 64.17$^{\dagger}$ \\
         \cmidrule(lr){2-9}
         % & SyncSwap-VQ & 12.1 & 0.903 & 9.527 & 1.462 & 7.87 & 89.9 & 81.66 \\
         & SwapTalk & 11.6 & 0.908 & 0.521 & 1.221 & 9.08 & 87.8 & 78.00 \\
         & SwapTalk (Extra Data) & \textbf{11.1} & \textbf{0.910} & 0.530 & \textbf{1.139} & 9.25 & \textbf{92.3} & \textbf{81.88} \\
        \midrule
        \multirow{8}{*}{Cross-Driven} & Wav2Lip~\cite{prajwal2020lip} & 27.4 & 0.811 & 0.417 & - & 7.87 & - & - \\
         & Wav2Lip-Restore & 14.6 & 0.877 & \textbf{0.533} & - & 7.63 & - & - \\
         & Sync-Swap & 12.7 & 0.899 & 0.483 & - & 7.52 & 84.2 & 77.83 \\
         & Sync-Swap-Restore & 13.0 & 0.885 & 0.528 & - & 7.24 & 83.8 & 77.44 \\
         & Swap-Sync & 11.6 & 0.900 & 0.484 & - & 8.62 & 82.6 & 76.63 \\
         & Swap-Restore-Sync & 11.8 & 0.888 & 0.521 & - & 8.57 & 80.4 & 75.99 \\
         \cmidrule(lr){2-9}
         % & SyncSwap-VQ & 11.9 & 0.903 & 0.522 & - & 7.62 & 88.1 & 80.83 \\
         & SwapTalk & 11.0 & 0.905 & 0.524 & - & 8.94 & 87.6 & 80.19 \\
         & SwapTalk (Extra Data) & \textbf{10.8} & \textbf{0.907} & 0.526 & - & \textbf{8.99} & \textbf{93.5} & \textbf{82.57} \\
        \bottomrule
    \end{tabular}
    \label{tab:main_expr_combined}
\end{table*}

\paragraph{Pre-training Data} During the pre-training phase of VQGAN, we utilize three public datasets: FFHQ~\cite{karras2019style}, CelebA-HQ~\cite{karras2017progressive}, and VFHQ~\cite{xie2022vfhq}. Additionally, we integrated approximately 143k images containing 33k different identities from private collections to enrich the model's learning materials. %The inclusion of these private datasets allows VQGAN to learn a more diverse representation of facial features.

\paragraph{Lip-Sync and Face Swapping Data} Following~\cite{bao2023wavsyncswap}, we employ the HDTF dataset~\cite{zhang2021flow} for training the lip-sync and face swapping modules. This dataset comprises 412 videos, each linked to a distinct speaker. We divide these videos into 5,586 ten-second clips, maintaining a frame rate of 25 frames per second. The dataset is split in a 9:1 ratio, employing an identity isolation strategy to ensure that speakers in the test set do not overlap with those in the training set. Employing this strategy is aimed at precisely assessing the model's ability to generalize to unseen identities. For the face swapping dataset, we consider each frame from the HDTF training videos as individual images of the same identity and randomly execute face swapping using Roop~\cite{roop}. This approach yields a total of 109k training data pairs (Source, Target, Res) for the face swapping module, with Res denoting the proxy model's outcome of transferring the source face onto the target.

To further enhance the generalization performance of the two modules, we expand the training dataset. For the face swapping module, additional data is sourced from the VQGAN pre-training dataset. We generate 216,000 training samples by creating random pairs and employing Roop to produce proxy-swapped faces. Meanwhile, the lip-sync module has been augmented with 85,000 10-second video clips, featuring 15,000 unique identities from private collections. These clips are processed like the HDTF dataset, ensuring consistency in data handling.

\subsection{Implementation Details}
During the data preprocessing phase, we adopt the same alignment method as FFHQ~\cite{karras2019style}, which involves using a 68 landmark detection technique to align faces and then crop and scale them to a resolution of 512 pixels. %To eliminate the influence of background noise, we further utilize MediaPipe~\cite{lugaresi2019mediapipe} to estimate the face mesh of the aligned faces and mask the background areas accordingly. 
During the pre-training of VQGAN, the hyperparameters of the loss function are set to $\lambda_{rec}=1.0, \lambda_{VQ}=1.0, \lambda_{id}=1.0$. Meanwhile, the batch size is set to 12 and the learning rate is adjusted to $4.5e-6$. Then, when training the face swapping module, we further set $\lambda_{res}=\lambda_{src}=1.0$ and adjust the batch size to 8 with a learning rate of $5e-5$. Additionally, when training the lip-sync module, we also set $\lambda_{sync}=0.3$, choose a larger batch size of 20, and increase the learning rate to $1e-4$. All these modules are implemented in the PyTorch with the Adam optimizer, trained on a server with 8 Nvidia RTX3090 GPUs.

\subsection{Experimental Results}
\paragraph{Evaluation Metrics} For the evaluation of video quality, we utilize two key metrics: the Fr\'{e}chet Inception Distance~\cite{heusel2017gans} (FID), which helps in understanding the similarity of generated videos to real ones, and the Structure SIMilarity~\cite{wang2004image} (SSIM), for assessing the preservation of structural information. To evaluate the clarity of the videos, we use the Cumulative Probability of Blur Detection (CPBD). For measuring the accuracy of lip synchronization, we rely on two metrics: the distance between specific landmarks around the mouth area (LandMark Distance, LMD) and the Lip Sync Error-Confidence~\cite{prajwal2020lip} (LSE-C). Furthermore, we use the ID Retrieve metric to evaluate the fidelity of face swapping and our proposed Consistency metric to assess identity consistency in the generated videos.

\paragraph{Description of Test Tasks} Throughout the testing phase, video clips are randomly chosen from the test set, assembling them into 112 (source clip, target clip) pairs for evaluation purposes. The testing tasks are divided into \textbf{self-driven} and \textbf{cross-drive} settings. In the self-driven setting, the model needs to swap the face in the target clip with that of the source clip, utilizing the audio from the target clip to drive the synthesized video's lip movements. This resembles the self-reconstruction task used in several lip-sync model evaluations~\cite{guan2023stylesync, wang2023lipformer, mukhopadhyay2024diff2lip}, where the ground truth of lip shapes in the same posture can be utilized to calculate the LMD metric. To minimize the risk of lip shape leakage in a self-driven setting, we always use the VQ-embedding of the first frame of the face-replaced target clip as a reference, inputting it into the lip-sync module. The task in a cross-driven setting involves replacing the face in the target clip with that in the source clip while using the audio from the source clip to drive the lip movements of the synthesized video. This scenario is closer to real-world application. However, due to the absence of ground truth for lip shapes in the same posture, the quality of lip synchronization can only be evaluated using the LSE-C metric.

\paragraph{Methods Used for Comparison} To validate the effectiveness of the proposed method, we design a series of experiments and primarily compare our method with the straightforward cascade models. To ensure comparability, we integrate the same lip-sync module in both our proposed method and other cascade-based approaches. Besides, the face swapping model used in these cascade-based methods is Roop. Additionally, we employ GFPGAN~\cite{wang2021gfpgan} to enhance the video quality. The specific cascade-based methods compared include:
\begin{itemize}
    \item \textbf{Sync-Swap}: Initiate by synchronizing lip movements, followed by face swapping via Roop.
    \item \textbf{Sync-Swap-Restore}: To address Roop's clarity deficits, we further refined facial details using GFPGAN.
    \item \textbf{Swap-Sync}: Inverts the Sync-Swap sequence, starting with face swapping, then refining lip synchronization with our module.
    \item \textbf{Swap-Restore-Sync}: After applying the Roop for face swapping, we use GFPGAN to enhance the frame clarity before proceeding to reconstruct the lip movements.
\end{itemize}
In addition to these cascading methods, we also compare our approach with the currently most popular lip-sync model, \textbf{Wav2Lip}~\cite{prajwal2020lip}. Given the lower resolution of videos generated by this model, we further applied GFPGAN to enhance the clarity in Wav2Lip's output, which is referred to as the \textbf{Wav2Lip-Restore} method. \textbf{WAVSYNCSWAP}~\cite{bao2023wavsyncswap}, which achieves two tasks end-to-end simultaneously, is also within our comparative scope. Our proposed methods include a model trained solely on the HDTF dataset, denoted as \textbf{SwapTalk}, and a model trained with additional data, referred to as \textbf{SwapTalk (Extra Data)}.

\begin{figure}[htbp]
    \centering
    \includegraphics[width=8.5cm]{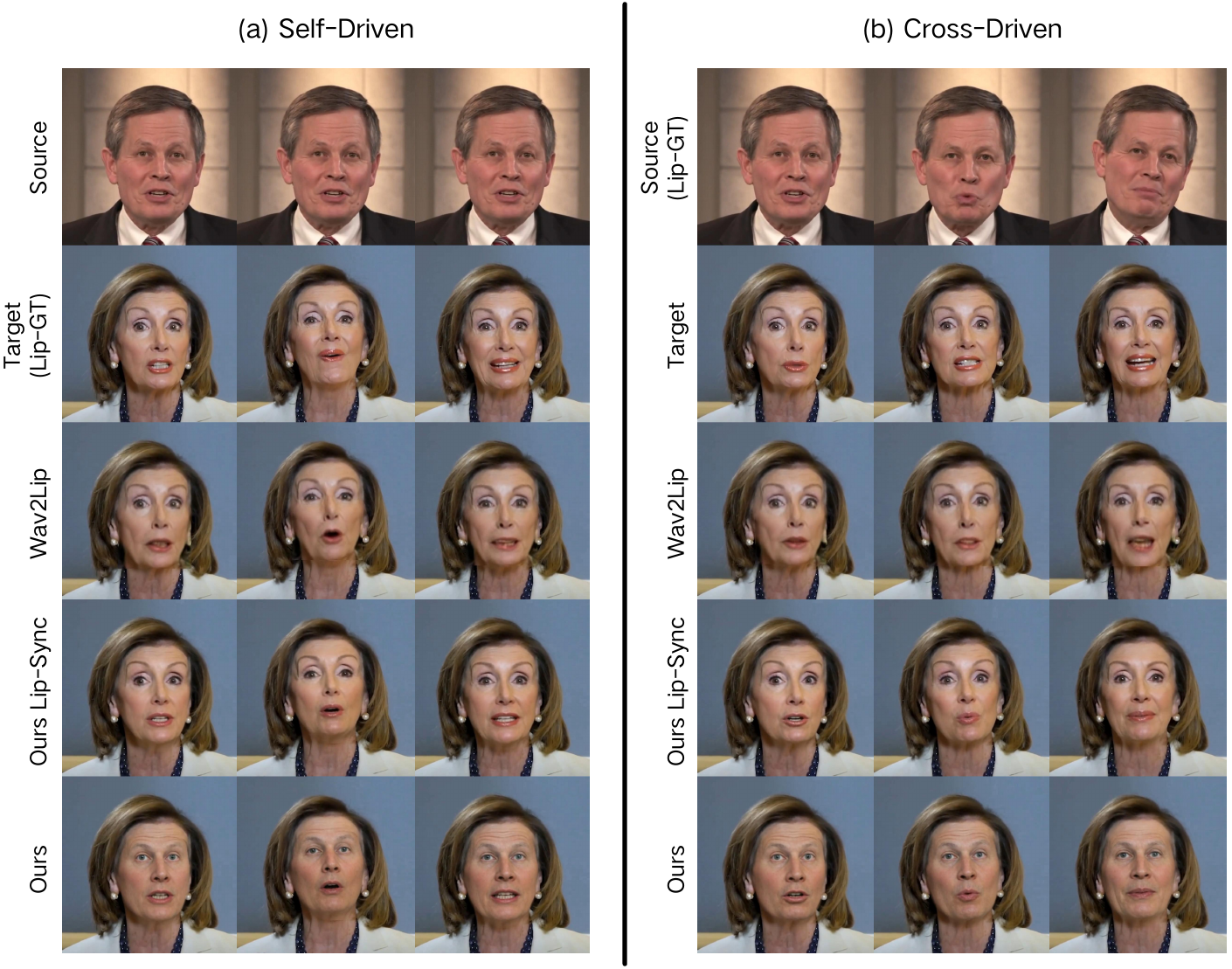}
    \caption{Our proposed method is compared with Wav2Lip under two settings: self-driven and cross-driven.
    % \textbf{Ours Lip-Sync} denotes the use of our trained lip-sync module without employing the face swapping module.
    }
    \label{fig:compare_wav2lip}
\end{figure}

\paragraph{Performance of Proposed Method}
Table~\ref{tab:main_expr_combined} presents the quantitative results of our model against other baseline models under two settings: self-driven and cross-driven. Without extra training data, our model surpasses other benchmarks in terms of image generation quality (as indicated by FID and SSIM metrics), lip synchronization accuracy (LMD or LSE-C), face swapping fidelity (ID Retrieve and Consistency). After incorporating additional training data, our model's performance further improved. Comparatively, while the enhancement in lip-sync quality is limited, face swapping fidelity and identity consistency improve significantly. In the self-driven task, Wav2Lip achieves higher scores on the LSE-C metric. We speculate this is because its training includes a discriminator that specifically focuses on supervising lip generation in the RGB space, which directly boosts its performance on this metric. Figure~\ref{fig:compare_wav2lip}(a) visually compares the results of Wav2Lip with our method. It also includes results from our lip-sync module alone (labeled as Ours Lip-Sync) to ensure a fair comparison. The visual results from the self-driven setting reveal that our models (Ours and Ours Lip-Sync) produce lip movements that align more closely with the actual ones, as evidenced by the lower values on the LMD metric. In the cross-driven setting, our model demonstrates superior performance in lip synchronization accuracy (LSE-C) compared to Wav2Lip, marking a distinct contrast from the self-driven scenario. Figure~\ref{fig:compare_wav2lip}(b) showcases the visual comparison between our method and Wav2Lip under this setting. We infer that the Wav2Lip model may rely more on the expressions and poses of the upper face rather than the audio information for predicting lip shapes, which could explain its inferior performance in handling arbitrary speech lip-sync tasks.

\begin{figure}[htbp]
    \centering
    \includegraphics[width=8.5cm]{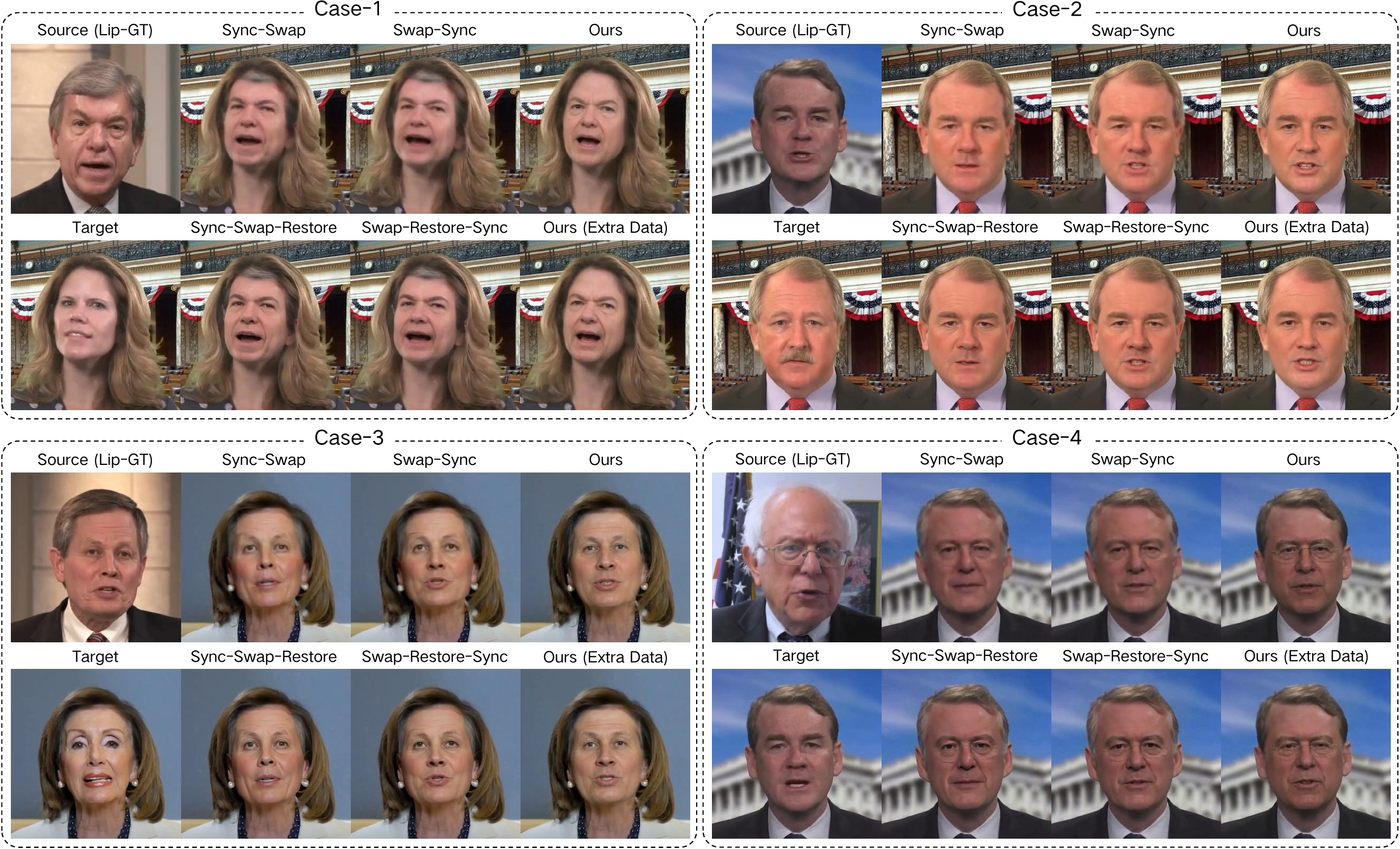}
    \caption{
    In the cross-driven scenario, a visual comparison of various cascade methods with the method we propose.
    }
\label{fig:cross-driven campare}
\end{figure}

In evaluating the cascade methods, our hypothesis that the two tasks might interfere with each other is confirmed. The comparisons in Table~\ref{tab:main_expr_combined}, across two different settings, indicate that Swap-Sync outperforms Sync-Swap in terms of lip synchronization quality (LMD or LSE-C). However, Sync-Swap exhibits superior performance in face swapping fidelity and identity consistency (measured by ID Retrieve and Consistency metrics). Figure~\ref{fig:cross-driven campare} visually contrasts our method with various cascade methods in the cross-driven scenario, showing that Sync-Swap falls short in lip performance compared to Swap-Sync. Furthermore, both numerical and visual results point out that integrating face restoration model (as seen in Sync-Swap-Restore and Swap-Restore-Sync schemes) significantly enhances image clarity (as measured by the CPBD metric) but at the expense of lip synchronization (LMD and LSE-C), face swapping fidelity, and video identity consistency (ID Retrieve and Consistency) decreased. Faces processed with image restoration technology often exhibit errors in facial texture, gaze, and lip detail accuracy, whereas our model demonstrates better performance in these aspects.

WAVSYNCSWAP attempts to address both tasks simultaneously in an end-to-end manner, yet its performance is unsatisfactory across image generation quality, lip synchronization, and face swapping fidelity. Our proposed method, which completes these two tasks within the same VQ-embedding latent space, has demonstrated a clear advantage over these baseline approaches.

\subsection{Ablation Studies}
To verify the effectiveness of our framework, we conduct a comprehensive analysis. The preliminary analysis focuses on the pre-training phase, exploring the impact of the spatial compression rate of VQGAN on model performance. For the face swapping module, we examined the importance of incorporating identity loss in improving model performance and the impact of using different backbones. For the lip-syncing module, in addition to analyzing the impact of different backbones, we also study the effectiveness of introducing a lip-sync expert in latent space. Finally, we explore the impact of the order in which two modules are cascaded in the VQ-embedding space during prediction.

\begin{table}[htbp]
    \centering
    \renewcommand{\arraystretch}{0.8}
    \caption{The impact of VQGAN with different spatial compression ratios on face swapping performance in HDTF dataset.}
    \begin{tabular}{cccc}
        \toprule
         Spatial Compress Rate & ID Retrieve$\uparrow$  & FID$\downarrow$\\
         \midrule
         8$\times$ & 85.63 & 15.6 \\
         16$\times$ & \textbf{94.29} & \textbf{9.5} \\
         \bottomrule
    \end{tabular}
    \label{tab:compare different vq compress ratio with face swapping}
\end{table}

\begin{figure}[htbp]
    \centering
    \includegraphics[width=8.5cm]{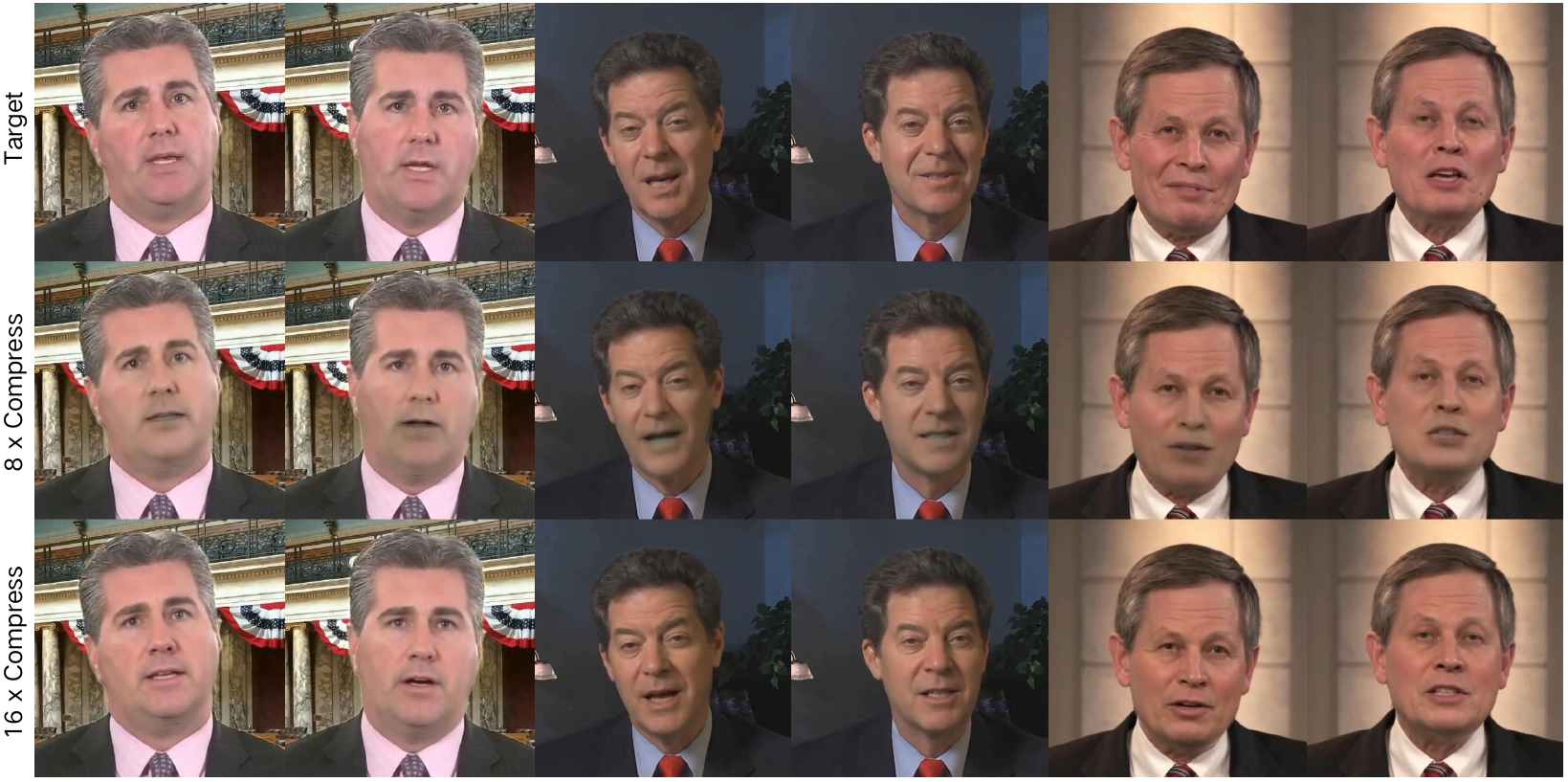}
    \caption{The impact of different VQ space compression ratios on the quality of generated videos.}
    \label{fig:compare_zip ratio}
\end{figure}

\paragraph{Exploration of VQGAN's Spatial Compression Rate}
Our pre-trained VQGAN model reduces the size of a $320\times 320$ face image by a factor of 16, compressing it into a $20\times 20$ VQ-embedding space. Given that the lip-sync and face swapping modules rely on this compact latent space, the space should maintain editability and high fidelity. We observe that the VQGAN's compression ratio significantly influences the performance of face swapping and lip-sync modules. We experiment with pre-training VQGAN at an 8$\times$ compression ratio, which expands the VQ-embedding space to $40\times 40$. Using this lower compression ratio, we train the face swapping and lip-sync modules with the identical model structure. Table~\ref{tab:compare different vq compress ratio with face swapping} and Table~\ref{tab:compare different vq compress ratio with lip-sync} show the performance comparison of two modules under two different compression ratios, and the results indicate that using a 16$\times$ compression ratio performs better than an 8$\times$ compression ratio on both tasks. We speculate that a lower compression ratio results in increased feature coupling in the latent space, complicating the learning process for both modules. Figure~\ref{fig:compare_zip ratio} displays the lip-sync results at different compression ratios; the 8$\times$ compression rate causes noticeable flickering and ghosting in the mouth area, whereas the 16$\times$ rate yields superior results.

\begin{table}[htbp]
    \centering
    \renewcommand{\arraystretch}{0.8}
    \caption{The impact of VQGAN with different spatial compression ratios on lip synchronization and video quality in HDTF dataset.}
    \begin{tabular}{cccc}
        \toprule
         Spatial Compress Rate & FID$\downarrow$ & SSIM$\uparrow$  & LMD$\downarrow$\\
         \midrule
         8$\times$ & 8.3 & 0.841 & 1.116 \\
         16$\times$ & \textbf{4.2} & \textbf{0.942} & \textbf{1.008} \\
         \bottomrule
    \end{tabular}
    \label{tab:compare different vq compress ratio with lip-sync}
\end{table}

\paragraph{Exploration of the Face Swapping} We explore the necessity of introducing two identity losses ($\mathcal{L}{id_res}$ and $\mathcal{L}{id_src}$) during the training process, as well as the impact of different backbone structures on the performance of the face swapping module. Table~\ref{tab:face swapper ablation study} reveals that without these identity losses, both the image generation quality and ID retrieval performance of the model are sub-optimal, indicating that relying solely on the proxy model's results is insufficient. Introducing $\mathcal{L}_{id\_src}$ significantly enhances the fidelity of face swapping, and further inclusion of $\mathcal{L}_{id\_res}$ notably improves the quality of the generated images. Additionally, we investigate face swapping models utilizing UNet and UNet from~\cite{Rombach_2022_CVPR} as backbones. Neither matches the performance of Transformers in terms of face swapping fidelity or image quality.

\begin{table}[htbp]
    \centering
    \renewcommand{\arraystretch}{0.8}
    \caption{Performance of different variants of face swapping modules.}
    \begin{tabular}{ccccc}
        \toprule
        $\mathcal{L}_{id\_src}$ & $\mathcal{L}_{id\_res}$ & Backbone & ID Retrieve$\uparrow$ & FID$\downarrow$ \\
        \midrule
         & & Transformer & 72.3 & 16.0 \\
         \checkmark &  & Transformer & 93.6 & 15.5 \\
         \checkmark & \checkmark  & Transformer & \textbf{94.3} & \textbf{9.5} \\
         \midrule
         \checkmark & \checkmark   &  UNet & 80.1 & 15.7 \\
         \checkmark & \checkmark   & UNet from ~\cite{Rombach_2022_CVPR} & 84.6 & 12.3 \\
         \bottomrule
    \end{tabular}
    \label{tab:face swapper ablation study}
\end{table}

\paragraph{Exploration of the Lip-Sync Module}
Similar to the experiments on face swapping module, we also investigate various backbone networks for the lip-sync module, as well as the impact of lip-sync expert supervision on module performance. In terms of network architecture, besides the UNet from~\cite{rombach2021highresolution} that we ultimately used, we also try a standard UNet and DiT~\cite{peebles2023scalable} as the backbone networks for our lip-sync module. When using the standard UNet, audio information is injected by concatenating the audio features along the channel dimension at the highest feature layer of the UNet and then upsampled to predict the lip shape VQ-embedding. With DiT, the audio information is treated as additional input, guided by Ada-Zero LN conditioned injection for generating the lip shape VQ-embedding. Table~\ref{tab:lip-sync ablation study} compares the impact of different backbone networks on the video quality (FID, SSIM) and lip-sync quality (LMD). The results show that using UNet from~\cite{rombach2021highresolution} as the backbone network for the lip-sync module is optimal. Additionally, we explore the effects of introducing lip-sync expert supervision in the VQ-embedding space. The LMD metrics from Table~\ref{tab:lip-sync ablation study} indicate that removing this supervision significantly degrades the quality of lip synchronization.

\begin{table}[htbp]
    \centering
    \renewcommand{\arraystretch}{0.8}
    \caption{The impact of different backbones.}
    
    \begin{tabular}{ccccc}
         \toprule
         Lip-Sync Expert & Backbone & FID$\downarrow$ & SSIM$\uparrow$  & LMD$\downarrow$ \\
         \midrule
         & UNet from~\cite{rombach2021highresolution}  & 4.3 & 0.940 & 1.381 \\
         \checkmark & UNet from~\cite{rombach2021highresolution} & \textbf{4.2} & \textbf{0.942} & \textbf{1.009} \\
         \checkmark & UNet  & 5.5 & 0.938 & 1.252 \\
         \checkmark & DiT  & 6.1 & 0.936 & 1.160\\
         \bottomrule
    \end{tabular}
    \label{tab:lip-sync ablation study}
\end{table}

\paragraph{Cascade Order of Modules in the VQ-embedding Space}
In the same VQ-embedding space, we investigate the cascading sequence of the face swapping and lip-sync modules. The SwapSync-VQ, which performs face swapping followed by lip-sync, significantly outperforms the SyncSwap-VQ, where lip-sync is conducted before face swapping, in terms of lip synchronization accuracy (as shown in Table~\ref{tab:cascade order in latent space}). Additionally, SwapSync-VQ matches SyncSwap-VQ closely in both face swapping fidelity and video identity consistency. Therefore, we ultimately selected the SwapSync-VQ for our methodology.

\begin{table}[htbp]
    \centering
    \renewcommand{\arraystretch}{0.8}
    \caption{The impact of different cascade orders.}
    
    \begin{tabular}{ccccccc}
        \toprule
         Cascade Order & FID$\downarrow$  & LSE-C$\uparrow$ & ID Retrieve$\uparrow$ & Consistency$\uparrow$\\
         \midrule
         SyncSwap-VQ & 11.9  & 7.62 & \textbf{88.1} & \textbf{80.83} \\
         SwapSync-VQ & \textbf{11.0} & \textbf{8.94} & 87.6 & 80.19 \\
         \bottomrule
    \end{tabular}
    \label{tab:cascade order in latent space}
\end{table}
\section{Conclusion}
In this study, we propose an innovative unified framework called SwapTalk for generating customized talking face videos. To address the issue of task interference and decreased video clarity in vanilla cascading existing models, we process face swapping and lip-sync tasks in the same editable and high-fidelity VQ-embedding space. The advantages of using VQ-embedding space include (1) reducing the computational cost of face swapping and lip-sync modules; and (2) leaving the high-resolution image generation task to VQGAN, reducing the model's learning difficulty. Additionally, we incorporate identity loss during the training phase of the face swapping module, which substantially enhances the model's capacity for generalizing to previously unseen identities. During the training of the lip-sync module, we employ supervision from a lip-sync expert within the VQ-embedding space, which significantly boosts the accuracy of lip synchronization. During the evaluation, we also introduce a novel metric specifically designed to quantify identity consistency in videos, further strengthening comprehensive assessment. On the public dataset HDTF, we conduct a comprehensive comparison between the SwapTalk framework and conventional cascade methods, end-to-end WAVSYNCSWAP model as well as Wav2Lip. From quantitative metrics and visual results, our proposed SwapTalk demonstrates significant advantages over other models in terms of video generation quality, lip synchronization accuracy, and face swapping fidelity.

\bibliographystyle{IEEEtran}
\bibliography{mybibfile}

\begin{thebibliography}{10}
\providecommand{\url}[1]{#1}
\csname url@samestyle\endcsname
\providecommand{\newblock}{\relax}
\providecommand{\bibinfo}[2]{#2}
\providecommand{\BIBentrySTDinterwordspacing}{\spaceskip=0pt\relax}
\providecommand{\BIBentryALTinterwordstretchfactor}{4}
\providecommand{\BIBentryALTinterwordspacing}{\spaceskip=\fontdimen2\font plus
\BIBentryALTinterwordstretchfactor\fontdimen3\font minus \fontdimen4\font\relax}
\providecommand{\BIBforeignlanguage}[2]{{%
\expandafter\ifx\csname l@#1\endcsname\relax
\typeout{** WARNING: IEEEtran.bst: No hyphenation pattern has been}%
\typeout{** loaded for the language `#1'. Using the pattern for}%
\typeout{** the default language instead.}%
\else
\language=\csname l@#1\endcsname
\fi
#2}}
\providecommand{\BIBdecl}{\relax}
\BIBdecl

\bibitem{Ki_2023_ICCV}
T.~Ki and D.~Min, ``Stylelipsync: Style-based personalized lip-sync video generation,'' in \emph{Proceedings of the IEEE/CVF International Conference on Computer Vision (ICCV)}, October 2023, pp. 22\,841--22\,850.

\bibitem{guan2023stylesync}
J.~Guan, Z.~Zhang, H.~Zhou, T.~Hu, K.~Wang, D.~He, H.~Feng, J.~Liu, E.~Ding, Z.~Liu \emph{et~al.}, ``Stylesync: High-fidelity generalized and personalized lip sync in style-based generator,'' in \emph{Proceedings of the IEEE/CVF Conference on Computer Vision and Pattern Recognition}, 2023, pp. 1505--1515.

\bibitem{gupta2023towards}
A.~Gupta, R.~Mukhopadhyay, S.~Balachandra, F.~F. Khan, V.~P. Namboodiri, and C.~Jawahar, ``Towards generating ultra-high resolution talking-face videos with lip synchronization,'' in \emph{Proceedings of the IEEE/CVF Winter Conference on Applications of Computer Vision}, 2023, pp. 5209--5218.

\bibitem{wang2023lipformer}
J.~Wang, K.~Zhao, S.~Zhang, Y.~Zhang, Y.~Shen, D.~Zhao, and J.~Zhou, ``Lipformer: High-fidelity and generalizable talking face generation with a pre-learned facial codebook,'' in \emph{Proceedings of the IEEE/CVF Conference on Computer Vision and Pattern Recognition}, 2023, pp. 13\,844--13\,853.

\bibitem{stypulkowski2024diffused}
M.~Stypu{\l}kowski, K.~Vougioukas, S.~He, M.~Zi{\k{e}}ba, S.~Petridis, and M.~Pantic, ``Diffused heads: Diffusion models beat gans on talking-face generation,'' in \emph{Proceedings of the IEEE/CVF Winter Conference on Applications of Computer Vision}, 2024, pp. 5091--5100.

\bibitem{Rombach_2022_CVPR}
R.~Rombach, A.~Blattmann, D.~Lorenz, P.~Esser, and B.~Ommer, ``High-resolution image synthesis with latent diffusion models,'' in \emph{Proceedings of the IEEE/CVF Conference on Computer Vision and Pattern Recognition (CVPR)}, June 2022, pp. 10\,684--10\,695.

\bibitem{Esser_2021_CVPR}
P.~Esser, R.~Rombach, and B.~Ommer, ``Taming transformers for high-resolution image synthesis,'' in \emph{Proceedings of the IEEE/CVF Conference on Computer Vision and Pattern Recognition (CVPR)}, June 2021, pp. 12\,873--12\,883.

\bibitem{vaswani2017attention}
A.~Vaswani, N.~Shazeer, N.~Parmar, J.~Uszkoreit, L.~Jones, A.~N. Gomez, {\L}.~Kaiser, and I.~Polosukhin, ``Attention is all you need,'' \emph{Advances in neural information processing systems}, vol.~30, 2017.

\bibitem{blanz2004exchanging}
V.~Blanz, K.~Scherbaum, T.~Vetter, and H.-P. Seidel, ``Exchanging faces in images,'' in \emph{Computer Graphics Forum}, vol.~23, no.~3.\hskip 1em plus 0.5em minus 0.4em\relax Wiley Online Library, 2004, pp. 669--676.

\bibitem{nirkin2018face}
Y.~Nirkin, I.~Masi, A.~T. Tuan, T.~Hassner, and G.~Medioni, ``On face segmentation, face swapping, and face perception,'' in \emph{2018 13th IEEE International Conference on Automatic Face \& Gesture Recognition (FG 2018)}.\hskip 1em plus 0.5em minus 0.4em\relax IEEE, 2018, pp. 98--105.

\bibitem{blanz2023morphable}
V.~Blanz and T.~Vetter, ``A morphable model for the synthesis of 3d faces,'' in \emph{Seminal Graphics Papers: Pushing the Boundaries, Volume 2}, 2023, pp. 157--164.

\bibitem{li2019faceshifter}
L.~Li, J.~Bao, H.~Yang, D.~Chen, and F.~Wen, ``Faceshifter: Towards high fidelity and occlusion aware face swapping,'' \emph{arXiv preprint arXiv:1912.13457}, 2019.

\bibitem{li2020advancing}
------, ``Advancing high fidelity identity swapping for forgery detection,'' in \emph{Proceedings of the IEEE/CVF conference on computer vision and pattern recognition}, 2020, pp. 5074--5083.

\bibitem{chen2020simswap}
R.~Chen, X.~Chen, B.~Ni, and Y.~Ge, ``Simswap: An efficient framework for high fidelity face swapping,'' in \emph{Proceedings of the 28th ACM International Conference on Multimedia}, 2020, pp. 2003--2011.

\bibitem{xu2022styleswap}
Z.~Xu, H.~Zhou, Z.~Hong, Z.~Liu, J.~Liu, Z.~Guo, J.~Han, J.~Liu, E.~Ding, and J.~Wang, ``Styleswap: Style-based generator empowers robust face swapping,'' in \emph{European Conference on Computer Vision}.\hskip 1em plus 0.5em minus 0.4em\relax Springer, 2022, pp. 661--677.

\bibitem{kim2022smooth}
J.~Kim, J.~Lee, and B.-T. Zhang, ``Smooth-swap: A simple enhancement for face-swapping with smoothness,'' in \emph{Proceedings of the IEEE/CVF Conference on Computer Vision and Pattern Recognition}, 2022, pp. 10\,779--10\,788.

\bibitem{li2021faceinpainter}
J.~Li, Z.~Li, J.~Cao, X.~Song, and R.~He, ``Faceinpainter: High fidelity face adaptation to heterogeneous domains,'' in \emph{Proceedings of the IEEE/CVF conference on computer vision and pattern recognition}, 2021, pp. 5089--5098.

\bibitem{wang2021hififace}
Y.~Wang, X.~Chen, J.~Zhu, W.~Chu, Y.~Tai, C.~Wang, J.~Li, Y.~Wu, F.~Huang, and R.~Ji, ``Hififace: 3d shape and semantic prior guided high fidelity face swapping,'' \emph{arXiv preprint arXiv:2106.09965}, 2021.

\bibitem{xu2022high}
Y.~Xu, B.~Deng, J.~Wang, Y.~Jing, J.~Pan, and S.~He, ``High-resolution face swapping via latent semantics disentanglement,'' in \emph{Proceedings of the IEEE/CVF conference on computer vision and pattern recognition}, 2022, pp. 7642--7651.

\bibitem{xu2022region}
C.~Xu, J.~Zhang, M.~Hua, Q.~He, Z.~Yi, and Y.~Liu, ``Region-aware face swapping,'' in \emph{Proceedings of the IEEE/CVF Conference on Computer Vision and Pattern Recognition}, 2022, pp. 7632--7641.

\bibitem{zhao2023diffswap}
W.~Zhao, Y.~Rao, W.~Shi, Z.~Liu, J.~Zhou, and J.~Lu, ``Diffswap: High-fidelity and controllable face swapping via 3d-aware masked diffusion,'' in \emph{Proceedings of the IEEE/CVF Conference on Computer Vision and Pattern Recognition}, 2023, pp. 8568--8577.

\bibitem{prajwal2020lip}
K.~Prajwal, R.~Mukhopadhyay, V.~P. Namboodiri, and C.~Jawahar, ``A lip sync expert is all you need for speech to lip generation in the wild,'' in \emph{Proceedings of the 28th ACM international conference on multimedia}, 2020, pp. 484--492.

\bibitem{alghamdi2022talking}
M.~M. Alghamdi, H.~Wang, A.~J. Bulpitt, and D.~C. Hogg, ``Talking head from speech audio using a pre-trained image generator,'' in \emph{Proceedings of the 30th ACM International Conference on Multimedia}, 2022, pp. 5228--5236.

\bibitem{karras2020analyzing}
T.~Karras, S.~Laine, M.~Aittala, J.~Hellsten, J.~Lehtinen, and T.~Aila, ``Analyzing and improving the image quality of stylegan,'' in \emph{Proceedings of the IEEE/CVF conference on computer vision and pattern recognition}, 2020, pp. 8110--8119.

\bibitem{yin2022styleheat}
F.~Yin, Y.~Zhang, X.~Cun, M.~Cao, Y.~Fan, X.~Wang, Q.~Bai, B.~Wu, J.~Wang, and Y.~Yang, ``Styleheat: One-shot high-resolution editable talking face generation via pre-trained stylegan,'' in \emph{European conference on computer vision}.\hskip 1em plus 0.5em minus 0.4em\relax Springer, 2022, pp. 85--101.

\bibitem{bao2023wavsyncswap}
W.~Bao, L.~Chen, C.~Zhou, S.~Yang, and Z.~Wu, ``Wavsyncswap: End-to-end portrait-customized audio-driven talking face generation,'' in \emph{ICASSP 2023-2023 IEEE International Conference on Acoustics, Speech and Signal Processing (ICASSP)}.\hskip 1em plus 0.5em minus 0.4em\relax IEEE, 2023, pp. 1--5.

\bibitem{wang2018high}
T.-C. Wang, M.-Y. Liu, J.-Y. Zhu, A.~Tao, J.~Kautz, and B.~Catanzaro, ``High-resolution image synthesis and semantic manipulation with conditional gans,'' in \emph{Proceedings of the IEEE conference on computer vision and pattern recognition}, 2018, pp. 8798--8807.

\bibitem{deng2019arcface}
J.~Deng, J.~Guo, N.~Xue, and S.~Zafeiriou, ``Arcface: Additive angular margin loss for deep face recognition,'' in \emph{Proceedings of the IEEE/CVF conference on computer vision and pattern recognition}, 2019, pp. 4690--4699.

\bibitem{hsu2021hubert}
W.-N. Hsu, B.~Bolte, Y.-H.~H. Tsai, K.~Lakhotia, R.~Salakhutdinov, and A.~Mohamed, ``Hubert: Self-supervised speech representation learning by masked prediction of hidden units,'' \emph{IEEE/ACM Transactions on Audio, Speech, and Language Processing}, vol.~29, pp. 3451--3460, 2021.

\bibitem{karras2019style}
T.~Karras, S.~Laine, and T.~Aila, ``A style-based generator architecture for generative adversarial networks,'' in \emph{Proceedings of the IEEE/CVF conference on computer vision and pattern recognition}, 2019, pp. 4401--4410.

\bibitem{karras2017progressive}
T.~Karras, T.~Aila, S.~Laine, and J.~Lehtinen, ``Progressive growing of gans for improved quality, stability, and variation,'' \emph{arXiv preprint arXiv:1710.10196}, 2017.

\bibitem{xie2022vfhq}
L.~Xie, X.~Wang, H.~Zhang, C.~Dong, and Y.~Shan, ``Vfhq: A high-quality dataset and benchmark for video face super-resolution,'' in \emph{Proceedings of the IEEE/CVF Conference on Computer Vision and Pattern Recognition}, 2022, pp. 657--666.

\bibitem{zhang2021flow}
Z.~Zhang, L.~Li, Y.~Ding, and C.~Fan, ``Flow-guided one-shot talking face generation with a high-resolution audio-visual dataset,'' in \emph{Proceedings of the IEEE/CVF Conference on Computer Vision and Pattern Recognition}, 2021, pp. 3661--3670.

\bibitem{roop}
s0md3v, ``roop,'' Source code available at GitHub repository, 2023, available online: \url{https://github.com/s0md3v/roop} (Accessed on 28 January 2024).

\bibitem{heusel2017gans}
M.~Heusel, H.~Ramsauer, T.~Unterthiner, B.~Nessler, and S.~Hochreiter, ``Gans trained by a two time-scale update rule converge to a local nash equilibrium,'' \emph{Advances in neural information processing systems}, vol.~30, 2017.

\bibitem{wang2004image}
Z.~Wang, A.~C. Bovik, H.~R. Sheikh, and E.~P. Simoncelli, ``Image quality assessment: from error visibility to structural similarity,'' \emph{IEEE transactions on image processing}, vol.~13, no.~4, pp. 600--612, 2004.

\bibitem{mukhopadhyay2024diff2lip}
S.~Mukhopadhyay, S.~Suri, R.~T. Gadde, and A.~Shrivastava, ``Diff2lip: Audio conditioned diffusion models for lip-synchronization,'' in \emph{Proceedings of the IEEE/CVF Winter Conference on Applications of Computer Vision}, 2024, pp. 5292--5302.

\bibitem{wang2021gfpgan}
X.~Wang, Y.~Li, H.~Zhang, and Y.~Shan, ``Towards real-world blind face restoration with generative facial prior,'' in \emph{The IEEE Conference on Computer Vision and Pattern Recognition (CVPR)}, 2021.

\bibitem{rombach2021highresolution}
R.~Rombach, A.~Blattmann, D.~Lorenz, P.~Esser, and B.~Ommer, ``High-resolution image synthesis with latent diffusion models,'' 2021.

\bibitem{peebles2023scalable}
W.~Peebles and S.~Xie, ``Scalable diffusion models with transformers,'' in \emph{Proceedings of the IEEE/CVF International Conference on Computer Vision}, 2023, pp. 4195--4205.

\end{thebibliography}

\end{document}